\documentclass[10pt,twocolumn,letterpaper]{article}

\usepackage{iccv}      
\usepackage[accsupp]{axessibility}  

\newcommand{\red}[1]{{\color{red}#1}}

\usepackage{makecell}

\definecolor{lightcyan}{rgb}{0.88,0.95,1}
\usepackage {adjustbox}

\usepackage{tcolorbox}
\usepackage{colortbl}
\usepackage{xcolor}
\usepackage{amsmath}
\usepackage{algorithmic}
\usepackage{algorithm}
\usepackage{bm}
\usepackage{booktabs}
\usepackage{comment}
\usepackage{multirow}
\usepackage{framed}

\usepackage{url}
\makeatletter
\newcommand{\figcaption}[1]{\def\@captype{figure}\caption{#1}}
\newcommand{\tblcaption}[1]{\def\@captype{table}\caption{#1}}
\makeatother

\newcount\K
\def\bla#1{
\K=0 \loop\ifnum\K<#1
{\textcolor[gray]{0.9}{{\it bla bla bla bla bla bla bla bla bla bla bla bla bla bla bla}}}
\advance\K by1\repeat
}

\def\paragraph#1{\noindent\textbf{#1.}}

 \usepackage{tcolorbox}
 \definecolor{Green}{RGB}{0,200,0}

\definecolor{iccvblue}{rgb}{0.21,0.49,0.74}
\usepackage[pagebackref,breaklinks,colorlinks,allcolors=iccvblue]{hyperref}

\def\paperID{786} 
\def\confName{ICCV}
\def\confYear{2025}

\title{CityNav: A Large-Scale Dataset for Real-World Aerial Navigation}

\author{
Jungdae Lee\textsuperscript{1}\thanks{Equal contribution} \quad
Taiki Miyanishi\textsuperscript{2,3}\footnotemark[1] \quad
Shuhei Kurita\textsuperscript{4,1} \quad
Koya Sakamoto\textsuperscript{2} \quad
Daichi Azuma\textsuperscript{2} \quad\\
Yutaka Matsuo\textsuperscript{2} \quad
Nakamasa Inoue\textsuperscript{1}
\\[2ex]
\textsuperscript{1}Institute of Science Tokyo \quad
\textsuperscript{2}The University of Tokyo \quad
\textsuperscript{3}ATR \quad\\
\textsuperscript{4}National Institute of Informatics \quad
\\[2ex]
{\tt\small https://water-cookie.github.io/city-nav-proj/}
}

\begin{document}
\maketitle
\begin{abstract}
Vision-and-language navigation (VLN) aims to develop agents capable of navigating in realistic environments. While recent cross-modal training approaches have significantly improved navigation performance in both indoor and outdoor scenarios, aerial navigation over real-world cities remains underexplored primarily due to limited datasets and the difficulty of integrating visual and geographic information.
To fill this gap, we introduce \textbf{CityNav}, the first large-scale real-world dataset for aerial VLN.
Our dataset consists of 32,637 human demonstration trajectories, each paired with a natural language description, covering 4.65 km$^2$ across two real cities: Cambridge and Birmingham.
In contrast to existing datasets composed of synthetic scenes such as AerialVLN, our dataset presents a unique challenge because agents must interpret spatial relationships between real-world landmarks and the navigation destination, making CityNav an essential benchmark for advancing aerial VLN.
Furthermore, as an initial step toward addressing this challenge, we provide a methodology of creating geographic semantic maps that can be used as an auxiliary modality input during navigation.
In our experiments, we compare performance of three representative aerial VLN agents (Seq2seq, CMA and AerialVLN models) and demonstrate that the semantic map representation significantly improves their navigation performance.
\end{abstract}

\begin{figure}[t]
\centering
\includegraphics[width=\linewidth]{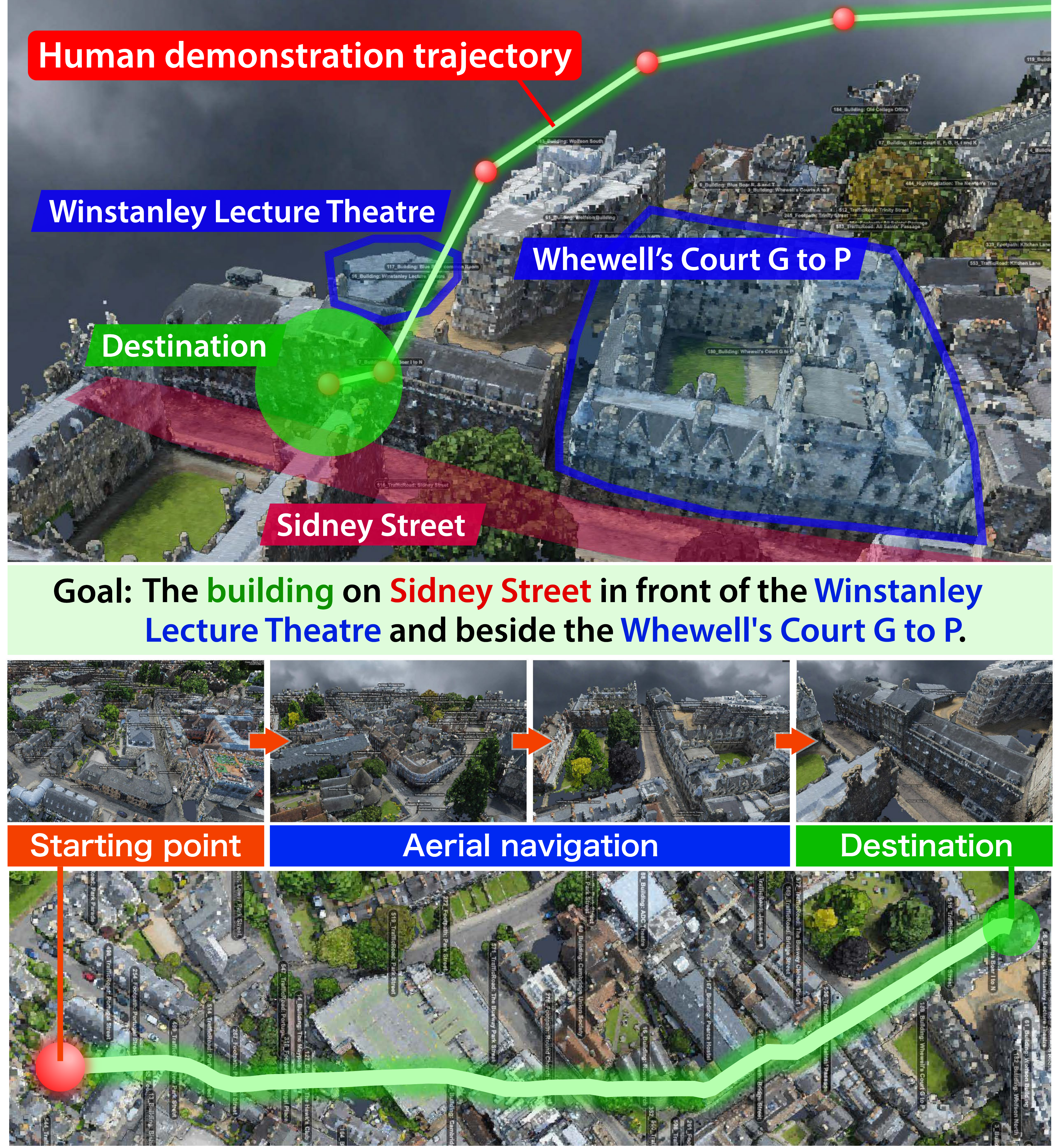}
\caption{
CityNav is a new aerial navigation dataset consisting of 32,637 human demonstration trajectories across real-world cities.
}
\label{fig:teaser}
\end{figure}

\section{Introduction}
Vision-and-language navigation (VLN) has emerged as a pivotal task in embodied artificial intelligence, in which an agent learns to navigate complex environments by following natural language description~\cite{nguyen-daume-iii-2019-help,Anderson2018Seq2seqCMA,krantz_vlnce_2020,zhou2024navgpt,Zhou2024NavGPT2UN}.
Over the past decade, significant progress has been made in developing VLN datasets and models across diverse environments, ranging from indoor house scenes~\cite{khanna2024goatbench,Ku2020rxr,Liu_2021_CVPR,Qi2020reverie,Zhu_2021_CVPR,eqamatterport} to outdoor urban scenes~\cite{Chen2019Touchdown,hermann2020learning,mirowski2018StreetLearn,vasudevan2021talk2nav,yang2024virl,wu2025metaurban}, 
with applications in robotics~\cite{Anderson2020SimtoRealTF,LM-Nav,zhang2024navid,long_2023_ICRA,Dai2024ThinkActAsk,long2024instructnav}.
\begin{table*}[t]
\def\cmark{\checkmark}
\small
\centering
\begin{tabular}{lcccccccc}
\toprule
\textbf{Dataset} &
\hspace{-5pt}\textbf{Real} & \textbf{Viewpoint} & \textbf{Environment} & \textbf{Trajectories} & \textbf{Instructions}& \textbf{Total length} & \textbf{Action space} & \textbf{Vocabulary}\\
\midrule
R2R~\cite{Anderson2018Seq2seqCMA} &\hspace{-5pt}{\cmark} & Ground & Indoor  & 7,189 & 21,567 & 71.9K & Graph & 3.1k \\
R$\times$R~\cite{Ku2020rxr}  &\hspace{-5pt}{\cmark} & Ground & Indoor  & 13,992  & 13,992 & 0.2M & Graph & 7.0k \\
CVDN~\cite{pmlr-v100-thomason20a} & \hspace{-5pt}{\cmark} & Ground & Indoor & 7,415 & 2,050 & -- & Graph & 4.4k \\
REVERIE~\cite{Qi2020reverie} &\hspace{-5pt}{\cmark} & Ground & Indoor  & 7,234 & 21,702 & 72.3K  & Graph & 1.6k \\ 
SOON~\cite{Zhu_2021_CVPR} & \hspace{-5pt}{\cmark} & Ground & Indoor  & 7,234 & 21,702 & 72.3K  & Graph & 1.6k \\
VLN-CE~\cite{krantz_vlnce_2020}  &\hspace{-5pt}{\cmark} & Ground & Indoor & 4,475 & 13,425 & 49.7K  & Graph & 4.3k \\
TouchDown~\cite{Chen2019Touchdown}  &\hspace{-5pt} {\cmark} & Ground & Outdoor & 9,326 & 9,326 & 2.9M  & Graph & 5.0k \\
Talk2Nav~\cite{vasudevan2021talk2nav} & \hspace{-5pt} {\cmark} &Ground & Outdoor & 10,714 & 10,714 & -- & Graph & 5.2k \\
\midrule
LANI~\cite{misra2018LANI} &\hspace{-5pt} & Aerial & Outdoor & 6,000 & 6,000 & 0.1M & 2 DoF & 2.3k \\
AVDN~\cite{fan2023ANDH}  &\hspace{-5pt} {\cmark} & Aerial & Outdoor & 3,064 & 6,269 & 0.9M & 3 DoF & 3.3k \\
AerialVLN~\cite{Liu2023AerialVLN} &\hspace{-5pt} & Aerial & Outdoor & 8,446 & 25,338 & 5.6M & 4 DoF & 4.5k \\
\textbf{CityNav (Ours)} & \hspace{-5pt} {\cmark} & Aerial & Outdoor & 32,637 & 32,637 & 17.8M & 4 DoF & 6.4k\\
\bottomrule
\end{tabular}
\caption{
Comparison of representative VLN datasets. Real: $\checkmark$ indicates datasets of a real-world environment.
}
\label{tab:related_datasets}
\end{table*}

To further broaden the scope of VLN, several recent studies have introduced aerial VLN datasets, which aim to facilitate autonomous navigation with unmanned aerial vehicles (UAVs).
For example, Fan et al.~\cite{fan2023ANDH} constructed the AVDN dataset, which provides 3k trajectories over 2D satellite images with human-human dialog.
Liu et al.~\cite{Liu2023AerialVLN} introduced the AerialVLN dataset, containing 9k trajectories in a 3D synthetic city environment.
This line of research opens new avenues for VLN research and serves as a critical foundation for real-world applications such as autonomous aerial delivery, disaster response, and environmental monitoring.
However, aerial VLN in real-world city environments remains underexplored due to the lack of high-fidelity 3D datasets with real-world complexity and the difficulty of integrating multimodal geographic and visual information for robust UAV navigation.

To address this limitation, we introduce \textbf{CityNav}, the first large-scale real-world 3D dataset for aerial {VLN}, which covers 4.65 km$^{2}$ across two real cities: Cambridge and Birmingham.
We collected 32,637 human demonstration trajectories, each representing a navigation path toward a designated goal object in 3D space.
To support this, we developed CityFlight, a realistic environment for flight simulation based on real-world 3D scan data.
Compared to existing datasets composed of synthetic 3D scenes such as AerialVLN, CityNav presents a unique challenge: agents must interpret spatial structure such as real-world landmarks and the navigation destination as shown in Figure~\ref{fig:teaser}.
To the best of our knowledge, CityNav is the largest aerial VLN dataset to date in terms of trajectory count, as evidenced by the comparison of representative VLN datasets in Table~\ref{tab:related_datasets}.

As CityNav introduces new challenges, we also provide a baseline approach as an initial step toward addressing them. Specifically, we introduce the geographic semantic map (GSM) representation from which agents can acquire geographic information including landmark locations. This representation can be integrated into existing VLN models as an auxiliary modality input, which in turn improves their navigation performance through training with human-demonstrated trajectories.
In experiments, we conduct extensive evaluation using three representative aerial VLN models: Seq2Seq~\cite{Anderson2018Seq2seqCMA}, cross-modal attention (CMA)~\cite{Anderson2018Seq2seqCMA}, and AerialVLN~\cite{Liu2023AerialVLN}, and demonstrate that the semantic map representation unlocks their ability to learn visual and geographic features for more effective navigation.
In summary, our main contributions are threefold:
\begin{enumerate}
\item[1)]
We introduce CityNav, a real-world aerial VLN dataset consisting of 32,637 human demonstration trajectories spanning 4.65 km$^{2}$ across two real cities. This dataset contains the largest number of aerial trajectories to date.

\item[2)]
We provide {the GSM representation} that enables existing VLN models to learn effective navigation paths from visual and geographic information.

\item[3)]
We conduct extensive experiments to evaluate three representative VLN models: Seq2seq, CMA and AerialVLN. We demonstrate that {the GSM representation} significantly improves their navigation performance.
\end{enumerate}

\begin{figure*}[t]
\centering
\includegraphics[width=1.0\linewidth]{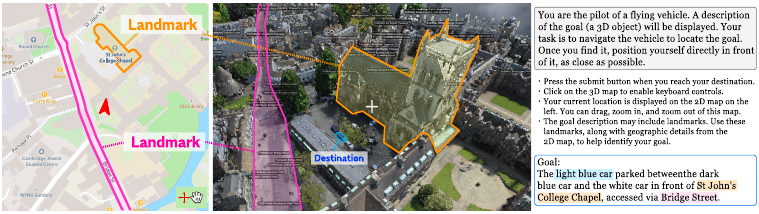}
\caption{
CityFlight is a 3D environment for flight simulation. Five actions, each mapped to a keyboard key, allow movement and rotation of the UAV. The 3D environment is synchronized with OpenStreetMap.
Human annotators are asked to navigate to the specified goal object within the 3D scene.
}
\label{fig:flight_simulator}
\end{figure*}

\section{Related Work}

High-fidelity 3D scanning technologies have opened new research avenues for embodied AI. Over the past decade, numerous datasets and tasks for 3D environments have been proposed, ranging from ground-level to aerial perspectives.

\paragraph{Ground-level Datasets}
Early efforts focused on computer vision tasks such as object detection and semantic segmentation in 3D indoor environments~\cite{scannetdata,ramakrishnan2021hm3d, habitat19iccv, xiazamirhe2018gibsonenv, dehghan2021arkitscenes, yeshwanthliu2023scannetpp, Matterport3D,armeni2016s3dis,ai2thor}. 
Building upon them, researchers have developed a variety of datasets and benchmarks for 3D vision-and-language tasks, including VLN~\cite{Anderson2018Seq2seqCMA, jain2019R4R, krantz_vlnce_2020, Ku2020rxr, rramrakhya2022, krantz2023iterative, wang2024grutopia}, vision-and-dialog navigation~\cite{nguyen-daume-iii-2019-help, pmlr-v100-thomason20a, huang2023embodied}, embodied referring expressions~\cite{Qi2020reverie, khanna2024goatbench, wang2024embodiedscan}, and embodied question answering~\cite{embodiedqa, eqamatterport, gordon2018iqa, mteqa, OpenEQA2023, Tan2021KnowledgebasedEQ}. 
Specifically, several studies have extended VLN to outdoor environments.
For example, TouchDown~\cite{Chen2019Touchdown} introduced navigation tasks in street-level environments, providing 9,326 human demonstration trajectories paired with natural language descriptions.
Talk2Nav~\cite{vasudevan2021talk2nav} collected 10,714 trajectories with verbal descriptions in an interactive environment based on Google Street View.

\paragraph{Aerial Datasets}
Aerial VLN is an emerging task focused on navigating flying vehicles such as UAVs in environments much larger than indoor spaces~\cite{misra2018LANI,fan2023ANDH,lam2018xview,gao2024embodiedcity,wang2025towards,zhao2025cityeqa,gao2025OpenFly}.
We highlight representative studies among these, starting with 
LANI~\cite{misra2018LANI}, the first dataset in this domain, which provides  6,000 pairs of natural language descriptions and trajectories in a small virtual environment.
AVDN~\cite{fan2023ANDH} collected 3,064 aerial navigation trajectories with human-to-agent dialog over 2D satellite images from xView~\cite{lam2018xview}.
Most recently, AerialVLN~\cite{Liu2023AerialVLN} provided 8,446 human demonstration trajectories in a 3D synthetic city environment, while proposing a navigation model that incorporates a look-ahead guidance mechanism into the cross-modal attention model~\cite{Anderson2018Seq2seqCMA}, achieving state-of-the-art aerial navigation performance.
Following this line of research, CityNav extends aerial VLN to large-scale real-world environments, posing the challenge of comprehending spatial relationships between real landmarks and designated goal object.

\section{CityNav Dataset}
\label{sec:citynav_dataset}

This section introduces the CityNav dataset, which provides the CityFlight environment for UAV-based aerial navigation over 4.65 km$^{2}$ across two real cities, along with a high-quality set of 32,637 human demonstration trajectories. 
A statistical comparison of CityNav with representative VLN datasets is shown in Table~\ref{tab:related_datasets}. To the best of our knowledge, CityNav offers the largest number of human demonstration trajectories to date for aerial VLN.

\subsection{CityFlight Environment}

To provide a realistic environment for flight simulation, we develop CityFlight on real-world 3D scan data. As shown in Figure~\ref{fig:flight_simulator},
this environment is synchronized with OpenStreetMap to allow agents to verify their location. UAVs can fly up to an altitude of 200 meters during navigation.

\paragraph{3D Scan Data}
We build CityFlight upon 3D point cloud data of Cambridge and Birmingham obtained from SensatUrban~\cite{hu2022SensatUrban} because this data is derived from real-world 3D scans, accurately representing actual cityscapes and thus enabling the effective utilization of real-world geographic information.

\paragraph{Action Space}
The position of an agent is represented as a 5D pose $\bm{p} = (x, y, z, \theta, \psi) \in \mathbb{R}^{5}$, where $(x, y, z)$ denotes the spatial coordinates, and $(\theta, \psi)$ denotes pitch and yaw.
The environment renders view images using the 3D data based on the given 5D pose.
The action space is then defined as $\mathcal{A}\hspace{-2pt}=\hspace{-2pt}\{\text{move-forward}, \text{turn-left}, \text{turn-right}, \text{ascend},$ $\hspace{2pt} \text{descend},\hspace{2pt} \text{stop}\}$.
The move-forward action moves the UAV forward by 5 meters in the direction it is facing.
The turn-left and turn-right actions rotate the UAV by 30 degrees counterclockwise and clockwise, respectively.
The ascend and descend actions move the UAV up and down by 2 meters, respectively.
The stop action halts the UAV's movement.
These actions are chosen for their effectiveness in aerial VLN~\cite{Liu2023AerialVLN}, as they balance simplicity and expressiveness.

\paragraph{OpenStreetMap}
To incorporate real-world geographic context, we provide functions for retrieving data from OpenStreetMap. Specifically, we offer 1) a function to convert between 3D coordinates in the environment and real-world 2D map coordinates, and 2) a function to obtain segments from landmark names.
This allows agents to use real-world spatial knowledge to improve navigation.

\paragraph{For the use of GNSS}
Our simulator can provide UAVs with absolute GNSS coordinates at any level of precision, which they may use during navigation. However, our CityNav setting assumes that the absolute coordinates of the goal are \emph{not} available, in line with existing VLN settings. Indeed, our primary goal is for UAVs to locate {a textually specified goal object}, compelling navigation models to interpret both geographical and visual cues following the descriptions. Consequently, we exclude any scenarios where the goal coordinates are known, as they fall outside the scope of existing VLN settings.

\paragraph{Implementation Details}
CityFlight is implemented with Potree~\cite{schutz2016potree}, an open-source WebGL-based point cloud renderer.
This enables 3D scene visualization in a web browser and thereby supports crowdsourced data collection via manual UAV operation.
During data collection, each of the six actions is assigned to a specific key press, and an additional rollback key is provided to undo mistakes.
OpenStreetMap is displayed alongside the 3D view, and both update synchronously with each movement action.

\subsection{Task Definition}
\label{sec:task_definition}
Within the CityFlight environment, 
the VLN task is to locate a goal object specified by a textual description, utilizing both visual and geographic information.

{\paragraph{Goal Description}}
To select candidate goal points, we use the CityRefer dataset~\cite{miyanishi_2023_NeurIPS}, which provides 35,196 natural language expressions for 5,800 objects in the SensatUrban dataset.
The goal of navigation is specified by 3D coordinates and described in natural language, either as a specific object (\textit{e.g.}, a car or a building) or as a particular location (\textit{e.g.}, a park or a parking lot).
Each description includes at least one landmark and specifies the goal object by describing its spatial relationships with surrounding objects.
The agent learns to navigate using visual and geographic observations from the environment.

\paragraph{Starting Point}
The starting point is randomly chosen within a 500-meter radius of the goal on the map. The altitude is randomly set between 100 and 150 meters.

\paragraph{Success Criteria} A navigation episode is deemed successful if the agent stops within a  20-meter spherical radius of the goal coordinates. 

\paragraph{Evaluation Metrics}
We use four metrics to evaluate navigation episodes: navigation error (NE), success rate (SR), oracle success rate (OSR), and success weighted by path length (SPL).
NE measures Euclidean distance from the agent's stopping point to the goal.
SR is the fraction of episodes where the agent meets the success criteria.
OSR is the fraction of episodes where at least one point during navigation meets the success criteria.
SPL is a metric where SR is weighted by the ratio of the optimal path length to the actual path taken, rewarding short navigation paths~\cite{pointgoalnav}.

\subsection{Human Demonstration Trajectories}

Human demonstration trajectories are collected via Amazon MTurk using the web interface of CityFlight. 
A total of 171 annotators who passed the qualification test contributed to the data collection, resulting in 32,637 high-quality description-trajectory pairs.
The distribution of goal object types was diverse, comprising 48.3\% buildings, 40.7\% cars, 7.4\% ground, and 3.6\% parking lots, each varying in size, shape, and color.

\paragraph{Annotation Interface}
In the CityFlight web interface, annotators are presented with a 3D view image, a 2D map and a textual description of the specified goal object.
Before annotation, we explained our research project to the annotators, including the objective of this study, the task definition,
and the fact that the data would be publicly released.
Upon submission, the trajectory from the starting point to the endpoint is recorded.
To help annotators assess their navigation quality, the distance to the goal is displayed after submission.

\begin{figure}[t]
\begin{minipage}{0.49\linewidth}
\centering
\includegraphics[width=0.97\linewidth]{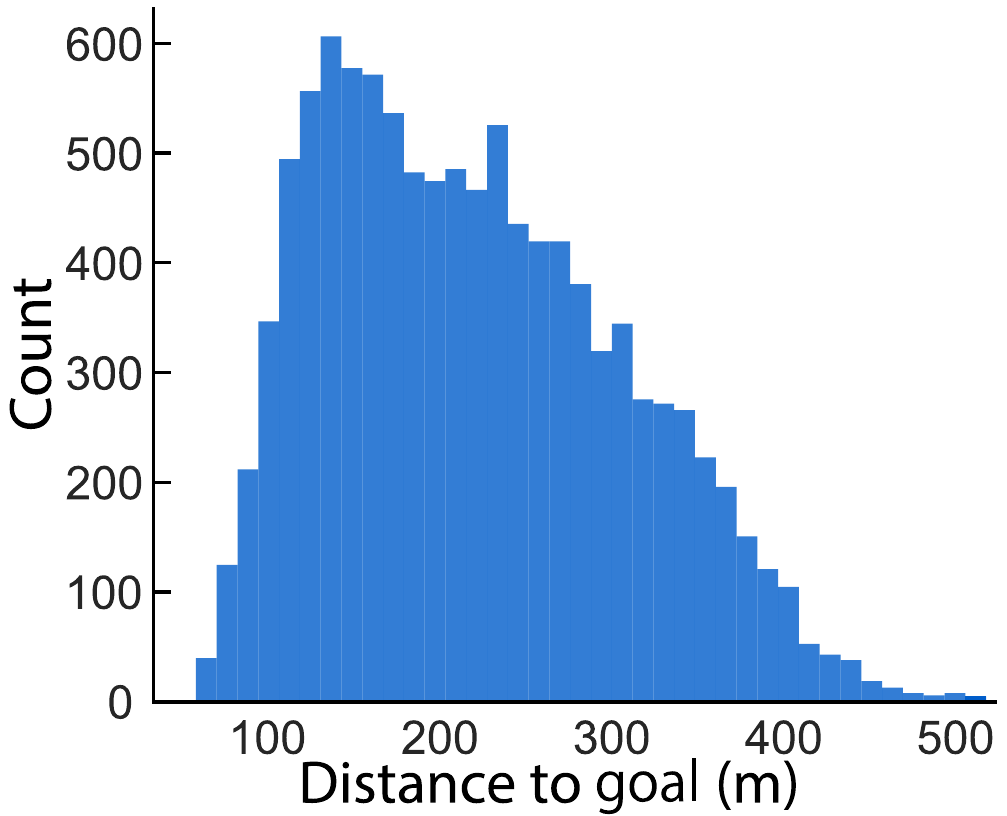}
\caption{Distance to goal}
\label{fig:dist_distance}
\end{minipage}
\begin{minipage}{0.49\linewidth}
\centering
\includegraphics[width=\linewidth]{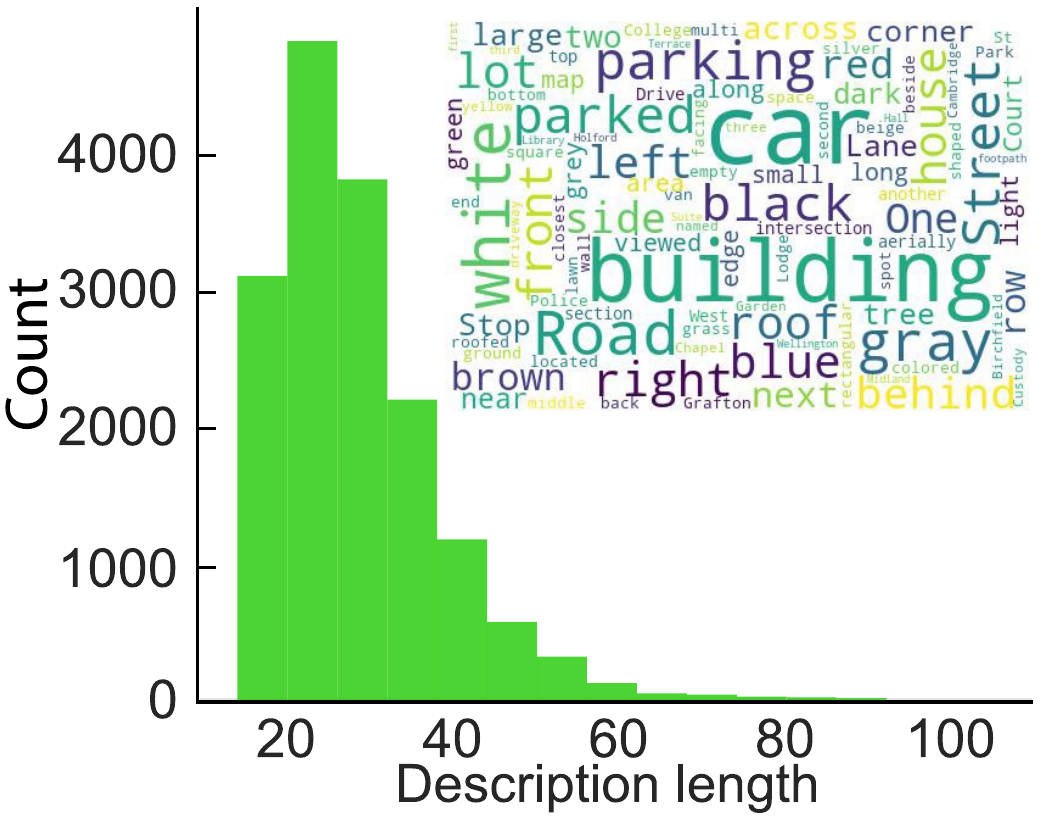}
\caption{Description length}
\label{fig:dist_instruction}
\end{minipage}
\end{figure}

\paragraph{Data Collection and Quality Control}
Human demonstration trajectories were collected via Amazon MTurk. To ensure high quality data, we conducted data collection in three stages: qualification, initial collection, and re-collection.
In the qualification stage, we screened annotators via navigation trials as the qualification tests to verify their ability to complete the task efficiently. Annotators who failed to reach their assigned goal or took more than twice the average completion time were excluded.
During the initial collection stage, trajectories were gathered for each goal. Among them, 18.4\% did not meet the success criteria and were excluded.
In the re-collection stage, annotators were asked to perform navigation for the excluded cases. Even after this step, trajectories for 7.2\% of goals still failed to meet the criteria and were removed from the dataset.
Through this rigorous process, we finalized 32,637 high-quality description-trajectory pairs for our dataset.

\paragraph{Additional Details}
Data collection required 711 total working hours at an hourly rate of \$12.83, totaling \$9,123. Overall, 171 annotators who passed the qualification contributed to the final dataset.
\begin{figure}[t]
\begin{minipage}{0.48\linewidth}
\centering
\includegraphics[width=\linewidth]{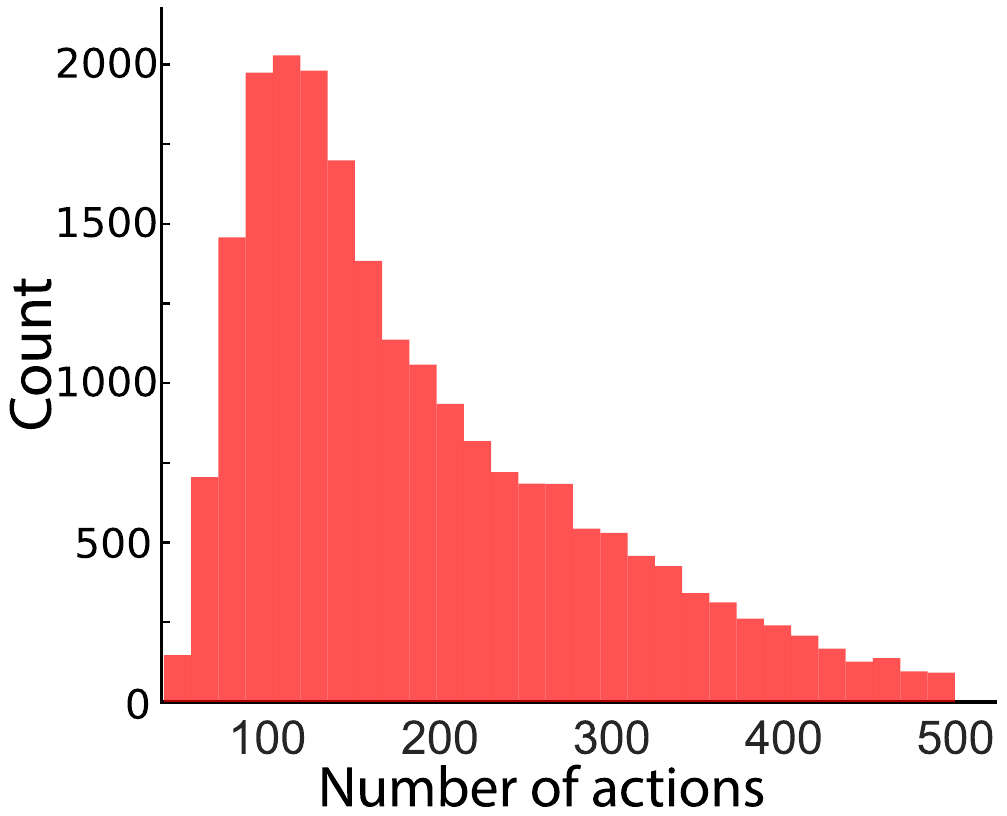}
\caption{Number of actions}
\label{fig:dist_num_actions}
\end{minipage}
\begin{minipage}{0.48\linewidth}
\centering
\includegraphics[width=\linewidth]{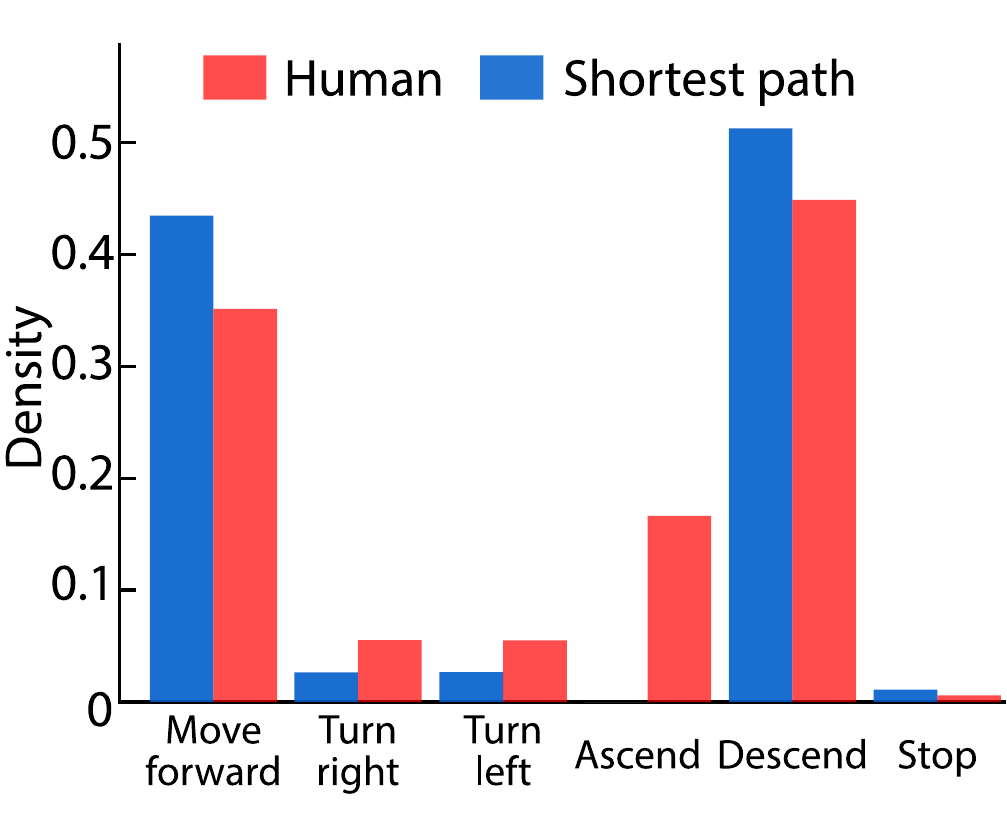}
\caption{Action distribution}
\label{fig:dist_action}
\end{minipage}
\end{figure}

\subsection{Dataset Statistics}
\label{sec:dataset_analysis}

\paragraph{Distance to Goal}
Figure~\ref{fig:dist_distance} shows the distribution of distances from the starting point to the goal. The distance varies from 50 meters to 500 meters, with a typical distance of around 200 meters. This variety allows for the evaluation of diverse navigation scenarios.

\paragraph{Description Lengths}
Figure~\ref{fig:dist_instruction} shows the distribution of description lengths in terms of the number of words, along with the word cloud representation of the word distribution. As shown, words indicating relative positions (\textit{e.g.}, left and right) as well as those representing visual information about the goal (\textit{e.g.}, colors and object parts) are frequently included. 
This shows that our dataset requires a comprehensive understanding of both visual and geographic information, opening new avenues for aerial VLN research.

\begin{figure}[t]
\centering
\includegraphics[width=\linewidth]{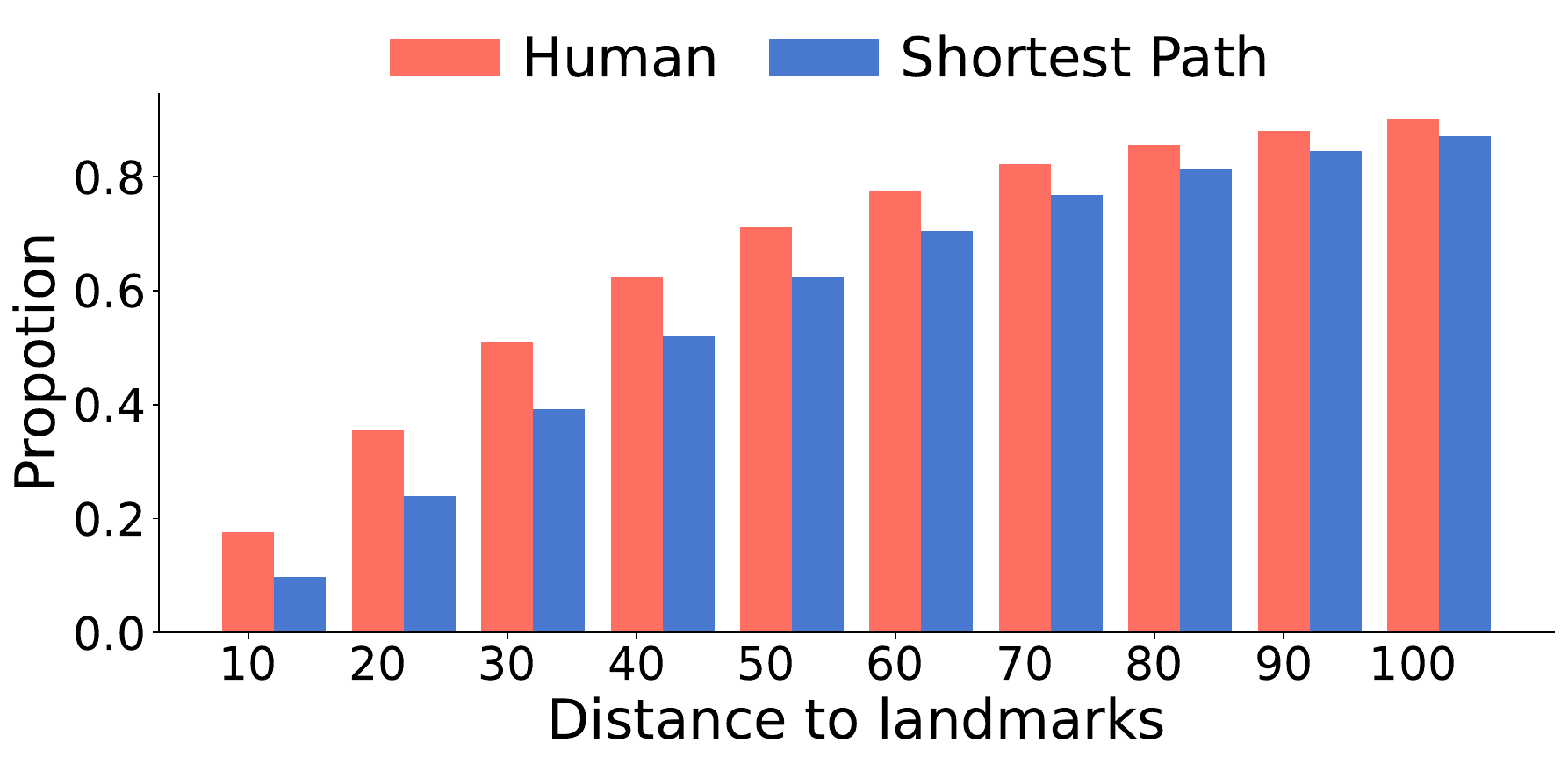}
\caption{Action proportion near landmarks.}
\label{fig:prob_hd_sp_landmark_dist}
\end{figure}

\paragraph{Number of Actions}
Figure~\ref{fig:dist_num_actions} shows the distribution of the number of actions required to reach the goal. The average number of actions is 240. Compared to existing datasets such as ADVN (49 actions) and AerialVLN (230 actions), our dataset is of comparable or greater trajectory length.

\paragraph{Action Distribution}
Figure~\ref{fig:dist_action} shows the distribution of actions compared to those taken during navigation along the shortest path. While the human annotators take some additional ascending movements to get a better view of surrounding objects, the overall distribution remains close to that of the shortest path.

\paragraph{Action Proportion Near Landmarks}
Figure~\ref{fig:prob_hd_sp_landmark_dist} shows the proportion of actions near landmarks.
For each trajectory, we extracted the landmark name from the associated description and determined whether the agent passed over the landmark polygon.
The results indicate that human-annotated trajectories tend to pass near landmarks more frequently compared to the shortest path.

\noindent Overall, the CityNav dataset covers a diverse range of navigation scenarios in real-world 3D environments, providing a rich resource for advancing research in aerial VLN.

\section{Models}
\label{sec:baselines}

We provide three baseline models for CityNav as an initial step toward addressing real-world aerial VLN. Specifically, we select three representative models: Seq2Seq~\cite{Anderson2018Seq2seqCMA}, CMA~\cite{Anderson2018Seq2seqCMA}, and AerialVLN~\cite{Liu2023AerialVLN}, because of their strong performance in aerial navigation.
Furthermore, to unlock their ability to leverage geographic information derived from 2D maps, we introduce the geographic semantic map representation that can be used as an auxiliary modality input to these models.

\subsection{Baseline Models}
\label{sec:baseline_models}

\paragraph{Seq2Seq~\cite{Anderson2018Seq2seqCMA}} This model employs a recurrent policy to predict the next action based on the current observation and navigation descriptions.
Specifically, it extracts hidden features using a GRU:
$\bm{h}_{t} = \mathrm{GRU}( [\bm{z}^{(t)}_{\text{RGB}}, \bm{z}^{(t)}_{\text{depth}}, \bm{z}_{\text{text}}], \bm{h}_{t-1})$, where
$\bm{z}^{(t)}_{\text{RGB}}$ is an RGB image representation extracted by a ResNet-50,
$\bm{z}^{(t)}_{\text{depth}}$ is a depth image representation extracted by another ResNet-50,
and $\bm{z}_{\text{text}}$ is a text representation of input descriptions extracted by an LSTM.
The action at time $t$ is then predicted from $\bm{h}_{t}$ through a learnable linear layer.

\begin{figure}[t]
\centering
\includegraphics[width=\linewidth]{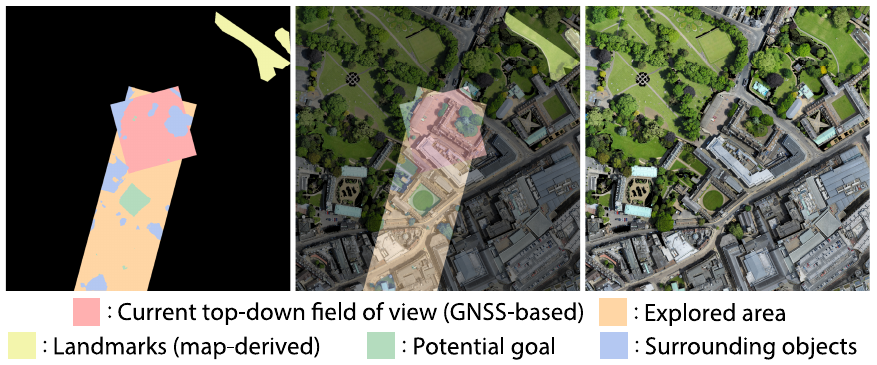}
\caption{Geographic semantic map (left) aligned with 3D scene (right).}
\label{fig:semantic_map}
\end{figure}

\begin{table*}[t]
\centering
\setlength{\tabcolsep}{4.7pt}
\begin{tabular}{lccccccccccccc}
\toprule
\multirow{2}{*}{\vspace{-6pt}Method} & \hspace{-1em}\multirow{2}{*}{\vspace{-6pt}GNSS}\hspace{-0.5em} & \multicolumn{4}{c}{Val-seen} & \multicolumn{4}{c}{Val-unseen} & \multicolumn{4}{c}{Test-unseen}\\
\cmidrule(lr){3-6} \cmidrule(lr){7-10} \cmidrule(lr){11-14}
&& NE$\downarrow$ & SR$\uparrow$ & OSR$\uparrow$ & SPL$\uparrow$ & NE$\downarrow$ & SR$\uparrow$ & OSR$\uparrow$ & SPL$\uparrow$ & NE$\downarrow$ & SR$\uparrow$ & OSR$\uparrow$ & SPL$\uparrow$ \\ \midrule 
Seq2Seq~\cite{Anderson2018Seq2seqCMA} && 257.1 & 1.81  &  7.89 & 1.58  & 317.4 & 0.79 & 8.82  & 0.61 & 245.3 & 1.50 & 8.34  & 1.30 \\
Seq2Seq\hspace{1pt}+\hspace{1pt}GSM &\checkmark& 58.5 & 8.43 & 17.31 & 7.28 & 78.6 & 5.13 & 10.90 & 4.65 & {98.1} & {3.81} & {13.92} & {2.79}\\
\midrule
CMA~\cite{Anderson2018Seq2seqCMA} && 240.8 & 0.95  &  9.42 & 0.92  & 268.8 & 0.65 & 7.86  & 0.63 & 252.6 & 0.82 & 9.70  & 0.79 \\
CMA\hspace{1pt}+\hspace{1pt}GSM
&\checkmark& 68.0 & 6.25 & 13.28 & 5.40 & 75.9 & 4.38 & 9.29 & 3.90 & 94.6 & 4.68 & 12.01 & 4.05
\\
\midrule
AerialVLN~\cite{Liu2023AerialVLN} & & {185.2} & {1.73} & {3.45} & {0.59} & {192.8} & {1.18} & {2.83} & {0.44} & {187.7} & {1.79} & {3.83} & {0.62} \\
AerialVLN\hspace{1pt}+\hspace{1pt}GSM & \checkmark & \textbf{56.6} & \textbf{10.16} & \textbf{22.20} & \textbf{7.89} & \textbf{72.7} & \textbf{6.35} & \textbf{15.24} & \textbf{5.06} & \textbf{85.1} & \textbf{6.72} & \textbf{18.21} & \textbf{5.16} \\
\midrule
Human && 9.1 & 89.31 & 96.40 & 60.17 & 9.4 & 88.39 & 95.54 & 62.66 & 9.8 & 87.86 & 95.29 & 57.04 \\
\bottomrule
\end{tabular}
\caption{Navigation performance comparison. Geographic semantic map representation (GSM) is incorporated into each baseline model.}
\label{tab:overall_results}
\end{table*}

\paragraph{Cross-Modal Attention (CMA)~\cite{Anderson2018Seq2seqCMA}} This model extends Seq2Seq by incorporating a bi-LSTM with cross-modal attention modules to enhance the integration of the RGB, depth, and text representations.
{This is a classical model originally proposed for indoor VLN, and its effectiveness in aerial VLN is reported in~\cite{Liu2023AerialVLN}.}

\paragraph{AerialVLN~\cite{Liu2023AerialVLN}}
This model {is a state-of-the-art model for aerial VLN}.
By introducing the look-ahead guidance to the CMA model, this model enables the navigation agent to learn to move toward several steps ahead during training.

\subsection{Geographic Semantic Map}
While the baseline models effectively integrate RGB, depth and text information for aerial navigation, they are not capable of leveraging real-world geographic information such as landmark locations. To address this limitation, we introduce the geographic semantic map (GSM), a simple yet effective representation that 
integrates map data retrieved from OpenStreetMap with the pose of agent. As shown in Figure~\ref{fig:semantic_map}, {the GSM} indicates informative areas such as landmarks and the explored area.

\paragraph{Categories}
{The GSM} consists of five categories: current field of view, explored area, landmarks, potential goals, and surrounding objects. These categories are selected because it is essential to understand the spatial relationships between the explored area and objects.
The current field of view and explored area are acquired from the GNSS coordinates. 
Landmarks are segments retrieved from OpenStreetMap.
Potential goals and surrounding objects are detected using an object detector (Grounding DINO~\cite{liu2023grounding}).
Before navigation begins, landmark and object names are extracted using a language model (GPT-3.5).
Binary masks aligned with the 2D map are generated for each category and used as a GSM.

\paragraph{Integration into VLN models}
The GSM $\bm{s}$ is encoded using a small convolutional neural network $E$ as $\bm{z}^{(t)}_{\text{map}} = E(\bm{s})$.
The GSM representation $\bm{z}^{(t)}_{\text{map}}$ can then be used as an auxiliary modality input.
For the three models in Section~\ref{sec:baseline_models},
$\bm{z}^{(t)}_{\text{map}}$ is integrated into the GRU module by appending it to the sequence $[\bm{z}^{(t)}_{\text{RGB}}, \bm{z}^{(t)}_{\text{depth}}]$.

\section{Experiments}
\label{sec:evaluation}
{We conduct extensive evaluations of VLN models on the CityNav dataset.
We first demonstrate that incorporating geographic semantic maps significantly improves the navigation performance across all baseline models. We then conduct a detailed analysis highlighting the importance of training with human demonstration trajectories and assessing model robustness in more challenging scenarios, including disaster simulations.}

\subsection{Experimental settings}
\label{sec:experimental_setup}

\paragraph{Dataset Split}
We divide the CityNav dataset into four distinct subsets: train, val-seen, val-unseen, and test-unseen.
{
The train and val-seen subsets, which contain 22,002 and 2,498 descriptions, respectively, share the same 24 scenes.
The val-unseen (2,826 descriptions, 4 scenes) and test-unseen (5,311 descriptions, 6 scenes) subsets
consist of distinct scenes not encountered during training, enabling evaluation of the model's generalizability to novel scenarios.
}

\paragraph{Evaluation Metrics} The four evaluation metrics are used: NE (meters), SR (\%), OSR (\%) and SPL (\%), as described in Section \ref{sec:task_definition}.

\paragraph{Implementation Details} 
The three {representative baseline models (Seq2Seq, CMA, and AerialVLN) are implemented with and without the geographic semantic map} as described in Section~\ref{sec:baseline_models}.
All models are trained on the train subset using the Adam optimizer for five epochs, with a learning rate of $1.5\times 10^{-3}$ and a batch size of 12.
The semantic map encoder $E$ consists of five convolutional layers with max-pooling and ReLU activations.
{
All models use the ResNet-50 encoders that are pre-trained on ImageNet~\cite{deng2009imagenet} and PointGoalNav~\cite{pointgoalnav} for the RGB and depth modalities, respectively.
}

\subsection{Experimental Results}
\noindent \textbf{Main Results.}
Table~\ref{tab:overall_results} summarizes the results on the three evaluation subsets.
As shown, the semantic map representation significantly improves the performance of all models. 
This is because it provides richer contextual information derived from a real-world 2D map, enabling better planning for more effective navigation.
Among the three models, AerialVLN performed the best, followed by CMA {in terms of navigation error in unseen environments}. This observation aligns with that of existing literature~\cite{Liu2023AerialVLN}, indicating that CityNav is consistent with prior benchmarks and effectively captures new challenges present in real-world aerial VLN.
When comparing the results of the val-seen and val-unseen subsets, navigation performance is consistently higher on val-seen across all metrics. This {indicates} that CityNav can serve as a valuable resource for developing VLN models that generalize to unseen city environments.

\paragraph{Human Evaluation}
We also report the results of human evaluation in Table~\ref{tab:overall_results}.
As shown, humans achieve a success rate of 87–90\% across all subsets. Since CityNav involves navigation tasks in extremely large environments, a significant performance gap remains between current VLN models and human navigators, highlighting a critical challenge for future research.

\begin{figure}[t]
    \centering
        \centering
        \includegraphics[width=0.75\linewidth]{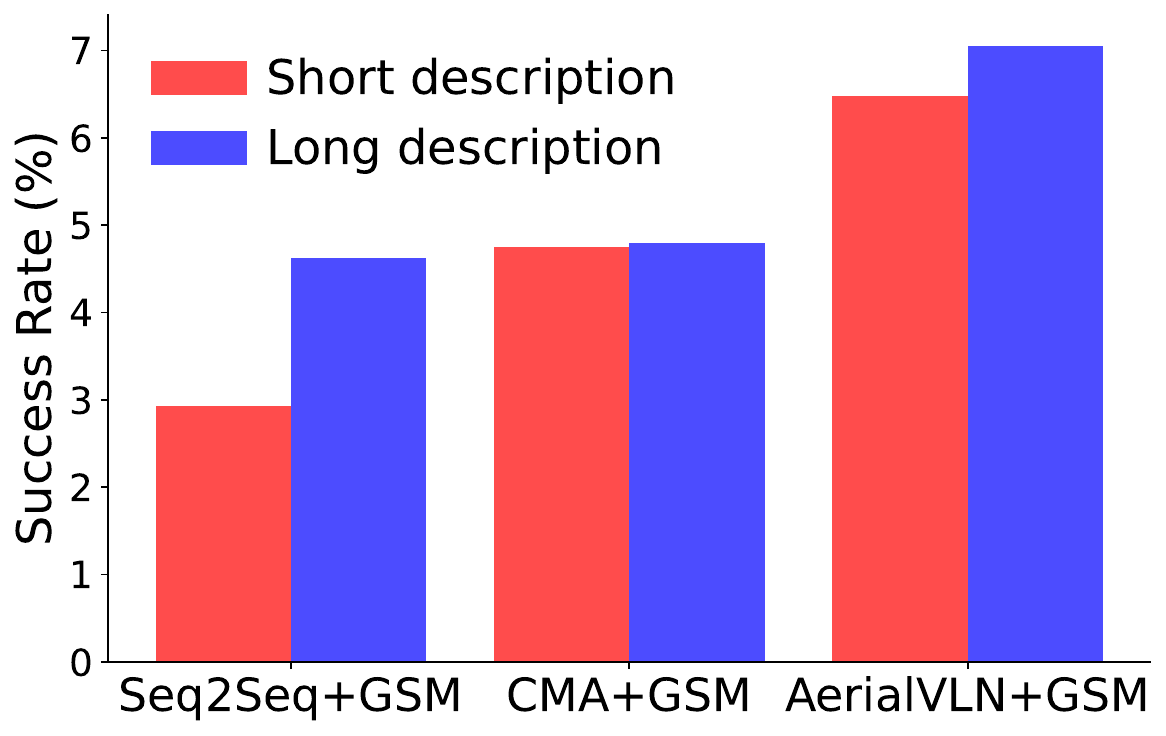}
    \caption{{Success rate comparison for short vs. long instructions (test-unseen). The categorization is based on the average instruction length and the number of landmarks.}}
    \label{fig:inst_land_analysis1}
\end{figure}

\begin{figure}[t]
    \centering
        \centering
        \includegraphics[width=0.75\linewidth]{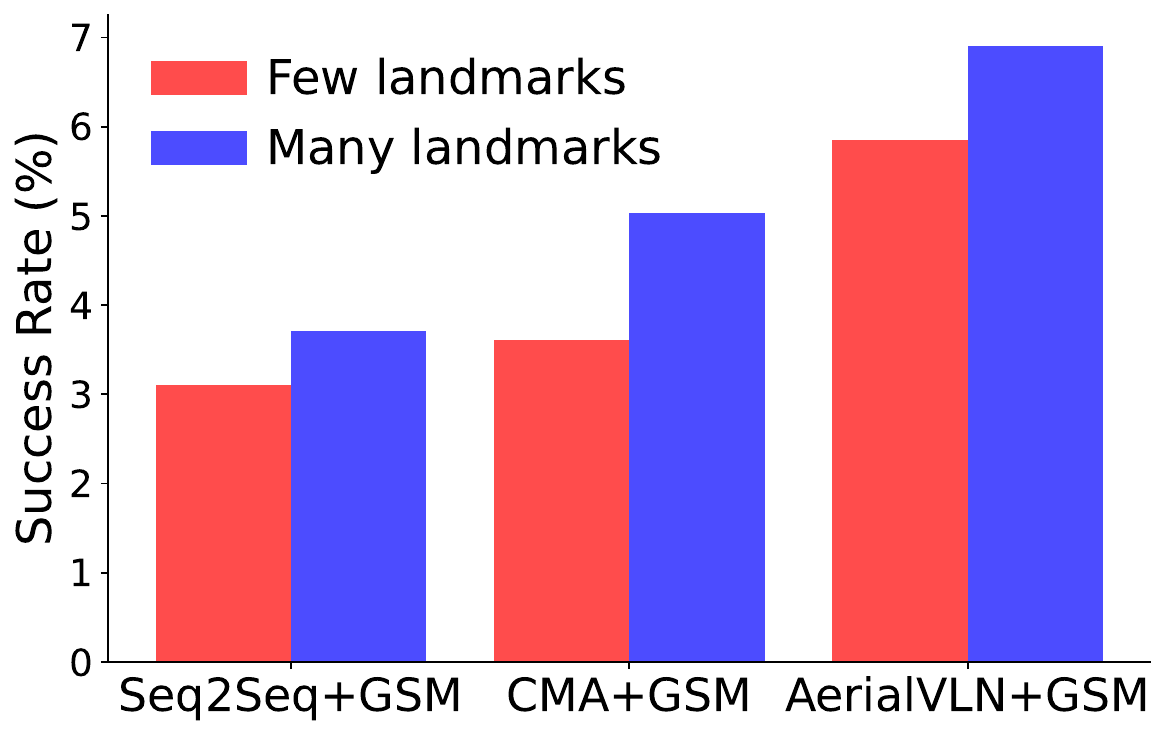}
    \caption{{Success rate comparison for few vs. many landmarks (test-unseen). The categorization is based on the average instruction length and the number of landmarks.}}
    \label{fig:inst_land_analysis2}
\end{figure}

\paragraph{Impact of description length and landmark density}
To better understand the effectiveness of GSM, we analyze its influence on navigation performance with respect to two critical factors: description length (short vs. long) and landmark density (few vs. many landmarks).
As shown in Figure~\ref{fig:inst_land_analysis1}, longer descriptions consistently lead to higher SR compared to shorter ones, suggesting that richer contextual details help models better align navigation actions with the goal object’s surroundings and relevant geographic cues.
Similarly, Figure~\ref{fig:inst_land_analysis2} indicates that higher landmark density boosts SR for all models, likely due to better utilization of landmarks.
These results highlight the importance of leveraging detailed {geographic} contextual information in descriptions.

\begin{table}[t]
\centering
\setlength{\tabcolsep}{2.2pt}
\begin{tabular}{lcccccc}
\toprule
Training data & Size & NE $\downarrow$ & SR $\uparrow$ & OSR $\uparrow$ & SPL $\uparrow$\\ 
\midrule
Shortest path & 8k & {99.1} & {4.43} & {13.49} & {4.36} \\
Human demonstrations & 8k 
& \textbf{92.8}
& \textbf{5.27}
& \textbf{15.59}
& \textbf{5.21} \\
\midrule
Shortest path & 22k & {95.1} & {4.96}& {15.22}& {4.85}\\
Human demonstrations & 22k 
& \textbf{85.1}
& \textbf{6.72}
& \textbf{18.21}
& \textbf{5.16} \\
\bottomrule
\end{tabular}
\caption{Comparison of training with shortest paths and human demonstration trajectories (test-unseen).}
\label{tab:necessity}
\end{table}
\begin{table}[t]
\centering
\setlength{\tabcolsep}{2.1pt}
\begin{tabular}{lcccccc}
\toprule
Training data  & Noise & NE $\downarrow$ & SR $\uparrow$ & OSR $\uparrow$ & SPL $\uparrow$\\ 
\midrule
Shortest path &  & 95.1 & 4.96 & 15.22 & 4.85 \\
Human demonstrations &  
& \textbf{85.1}
& \textbf{6.72}
& \textbf{18.21}
& \textbf{5.16} \\
\midrule
Shortest path & \checkmark & 123.1 & 2.37 & 6.88 & 2.35 \\
Human demonstrations & \checkmark
& \textbf{95.0}
& \textbf{4.92}
& \textbf{12.77}
& \textbf{4.79} \\
\bottomrule
\end{tabular}
\vspace{-2pt}
\caption{Noise robustness evaluation results (test-unseen).
}
\label{tab:pose_noise}
\vspace{-3pt}
\end{table}

\paragraph{Necessity of human demonstrations}
The effectiveness of training with shortest-path trajectories has been reported in {some} previous studies {for indoor environments}~\cite{objgoalchaplot2020object, embodiedqa, eqamatterport}, raising the question of whether human demonstration trajectories are necessary for training.
In response to this question, we compare the AerialVLN+GSM models trained on shortest-path trajectories and human demonstration trajectories in Table~\ref{tab:necessity}.
We observe that models trained on human demonstration trajectories achieve higher performance.
This is because training on shortest-path trajectories fail to learn the relative positions of the goal and other objects, as landmarks and surrounding objects may not be observed along the paths.
{Notably, when the amount of training data is increased, the performance improvement is more substantial for the model trained on human demonstration trajectories, showing a clear margin over that trained on shortest-path trajectories. This indicates that CityNav involves complexities that cannot be fully addressed by relying solely on simple shortest paths.}

{
Moreover, training on human demonstration trajectories leads to superior robustness. To demonstrate this, we conduct robustness evaluation experiments by adding Gaussian noise ($\pm$100 meters) to the agent's position during the testing phase. As shown in Table~\ref{tab:pose_noise},
the performance gap between models trained on human demonstration trajectories and shortest paths widens significantly (\textit{e.g.}, from 10.0 to 28.1 in NE). This robustness is due to the greater trajectory diversity in human demonstration data.
}

\begin{table}[t]
\centering
\setlength{\tabcolsep}{4pt}
\begin{tabular}{lccccc} 
\toprule
Method & NE $\downarrow$ & SR $\uparrow$ & OSR $\uparrow$ & SPL $\uparrow$\\ 
\midrule
AerialVLN+GSM & \textbf{85.1} & \textbf{6.72} & \textbf{18.21} & \textbf{5.16}\\
w/o landmarks & {190.7} & {0.60} & {6.94} & {0.56}\\
w/o potential destination & {92.8} & {3.97} & {13.08} & {3.86}\\
w/o surroundings objects & {87.5} & {5.17} & {15.16} & {5.10}\\
\bottomrule
\end{tabular}
\caption{Ablation study (test-unseen).}
\label{tab:ablation_results}
\end{table}

\begin{figure*}[t]
    \centering \includegraphics[width=0.92\textwidth]{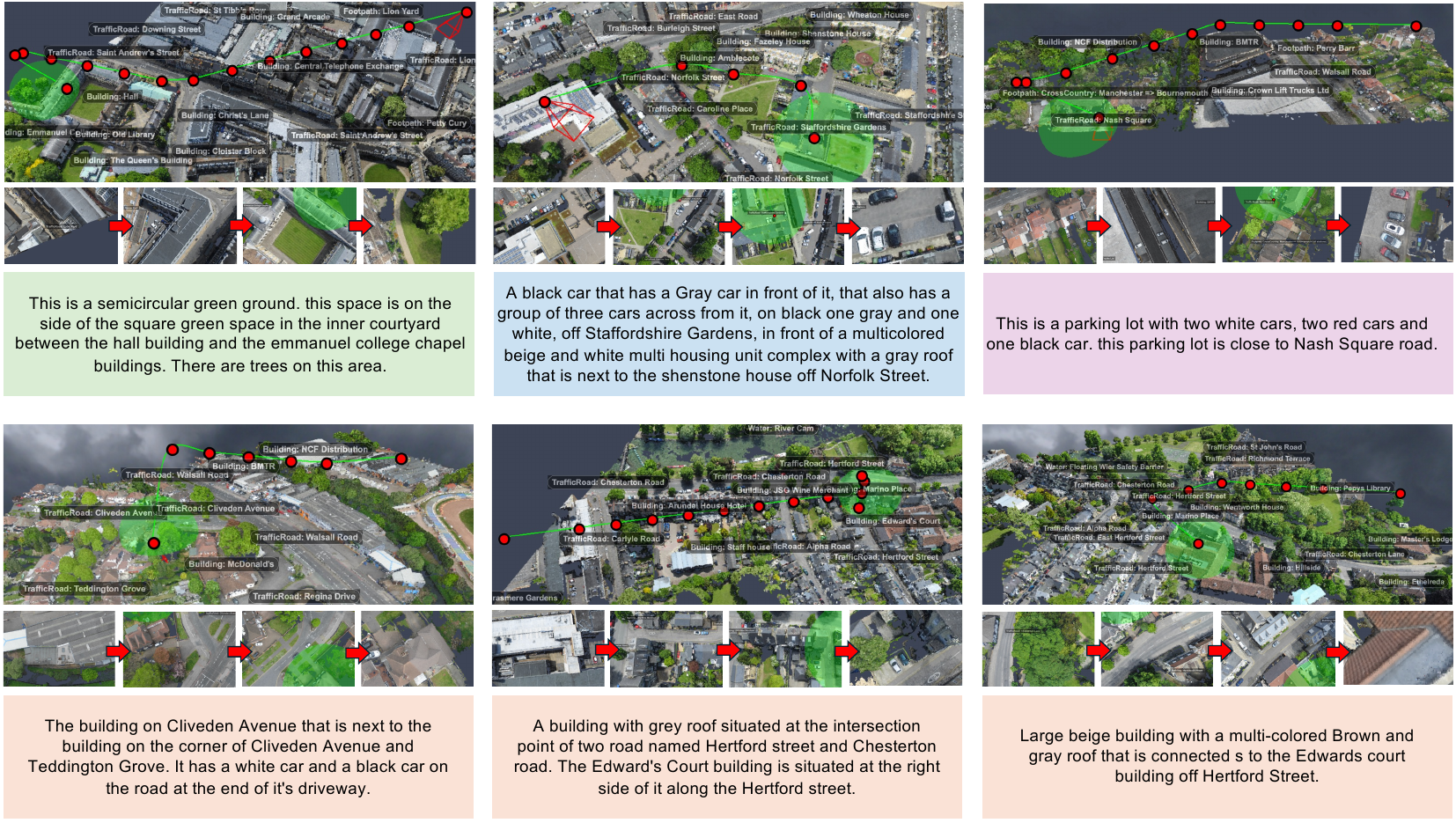} 
    \vspace{-4pt}
    \caption{Qualitative examples.}
    \label{fig:qualitaive}
    \vspace{-8pt}
\end{figure*}

\paragraph{Ablation study}
Table~\ref{tab:ablation_results} shows the results of the ablation study on the {GSM} representation. As shown, each component contributes to improving performance. We also observe that removing landmark information leads to a significant drop in performance. This indicates that learning both visual and geographical information is essential.

\paragraph{Qualitative examples}
Figure~\ref{fig:qualitaive} shows qualitative examples.
These examples demonstrate how integrating geographic data with navigation models enhances spatial awareness and target localization. 
By leveraging geographic cues, the agent effectively aligns its movements with described landmarks, refines its trajectory based on environmental context, and adapts to complex descriptions. 
In the top-left example, the aerial agent navigates through an urban environment by recognizing key landmarks such as ``Hall'' and ``Emmanuel College Chapel,'' the agent refines its route and successfully identifies the semicircular green ground described in the accompanying text ``trees on this area''.
These findings highlight the potential of combining linguistic and geographic information to improve navigation efficiency in real-world scenarios.

\section{Conclusion}
\label{sec:conclusion}
In this paper, we introduced CityNav, a new dataset for real-world aerial navigation that covers 4.65 km$^{2}$ across Cambridge and Birmingham.
We collected 32,637 high-quality human demonstration trajectories, making CityNav the largest aerial VLN dataset to date. As an initial step toward real-world aerial VLN, we provided three baseline models and introduced a semantic map representation. In our experiments, we evaluated the performance of these models and demonstrated that the semantic map representation substantially improves their performance. We also investigated a challenging scenario simulating a flooding disaster to assess the robustness of each model. 
Finally, we discuss limitations and future work.

\paragraph{Limitations}
Although CityNav spans large areas of two real cities, its coverage remains limited when compared to cities worldwide. Expanding the dataset to cover more urban regions would require reducing the cost of conducting 3D scans of outdoor environments.

\paragraph{Future research directions}
A promising avenue for future work involves exploring multi-agent collaboration when searching for a goal object in large 3D environments.
In this work, annotators performed navigation tasks independently, primarily because coordinating multiple annotators on Amazon MTurk is challenging. However, real-world scenarios often involve collaboration among multiple individuals, making a multi-agent dataset an appealing objective. Another interesting direction is integrating aerial, ground-level, and indoor navigation within a unified framework. Achieving this goal is difficult due to the substantial differences in the granularity of 3D scans required for these three domains. We believe CityNav lays an important foundation for advancing VLN in real-world applications and for contributing to these broader research efforts.

\noindent {\small \textbf{Acknowledgments.}
This work was supported by JST PRESTO (Grant Number JPMJPR22P8) and JSPS KAKENHI (Grant Numbers 25K03177 and 25K03135), Japan.
This work used the TSUBAME4.0 supercomputer at Institute of Science Tokyo.}

{
    \small
    \bibliographystyle{ieeenat_fullname}
    \bibliography{egbib}
}

\appendix

\clearpage
\begin{center}
{\rmfamily {\Large APPENDIX}}
\end{center}

\setcounter{section}{0}
This is appendix for the paper: {\it CityNav: A Large-Scale Dataset for Real-World Aerial Navigation}.
We present additional details of the data collection interface, dataset statistics, models, and experimental results.

\section{Data Collection Interface}
\begin{figure}[h]
    \centering
    \fbox{\includegraphics[width=0.46\textwidth]{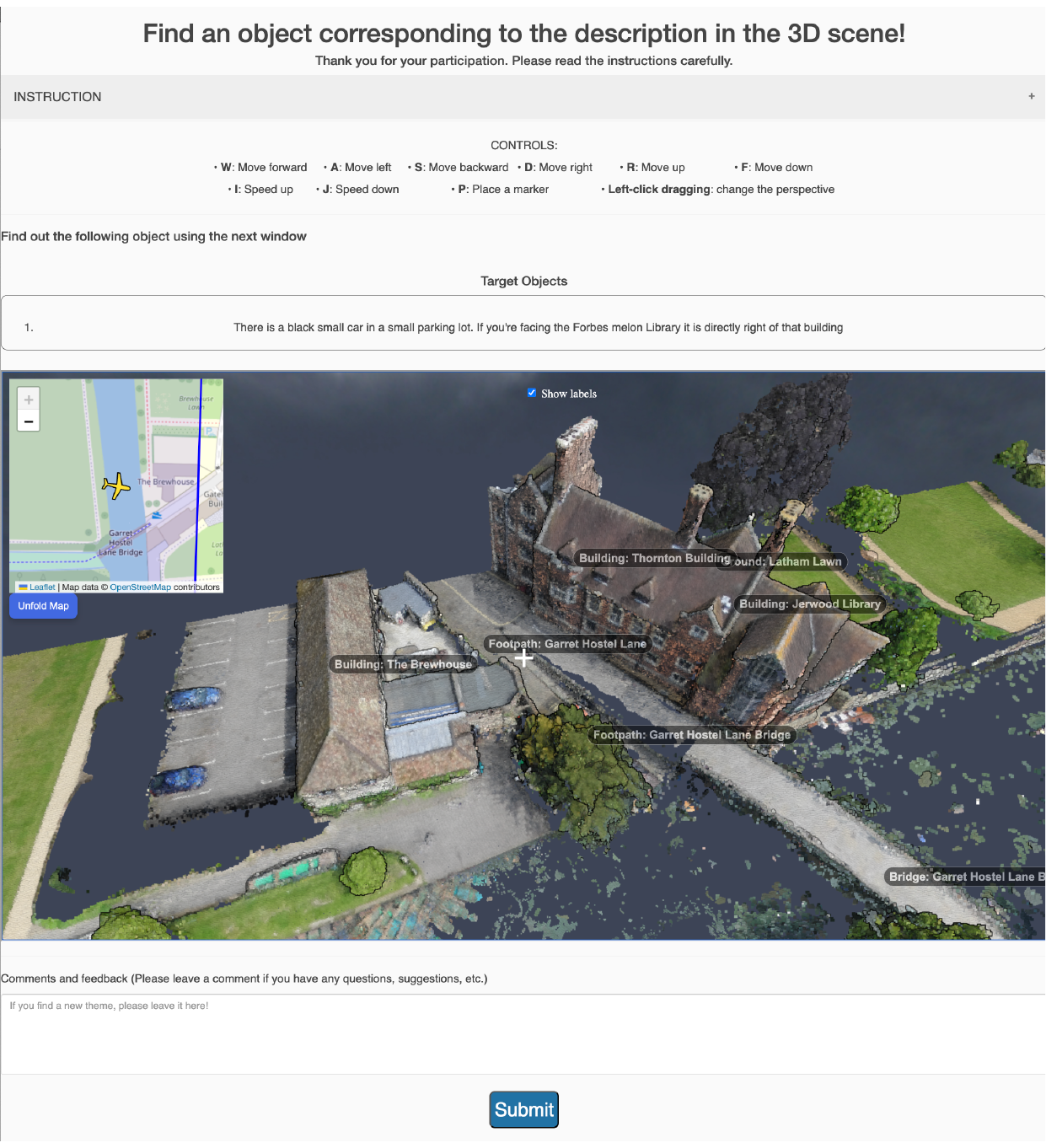}}
    \caption{
    \textbf{Data collection interface}. 
    Full screenshot of web interface for collecting human demonstration trajectories for the CityNav dataset. 
    }
\label{fig:web_interface_large}
    \vspace{-5pt}
\end{figure}

We developed the data collection website using the Amazon Mechanical Turk platform. 
Figure~\ref{fig:web_interface_large} displays a full screenshot of the web interface, enabling users to operate an aerial agent within the CityFlight environment.

\section{Dataset Statistics}
\begin{figure}[t]
    \centering
    \includegraphics[width=0.48\textwidth]{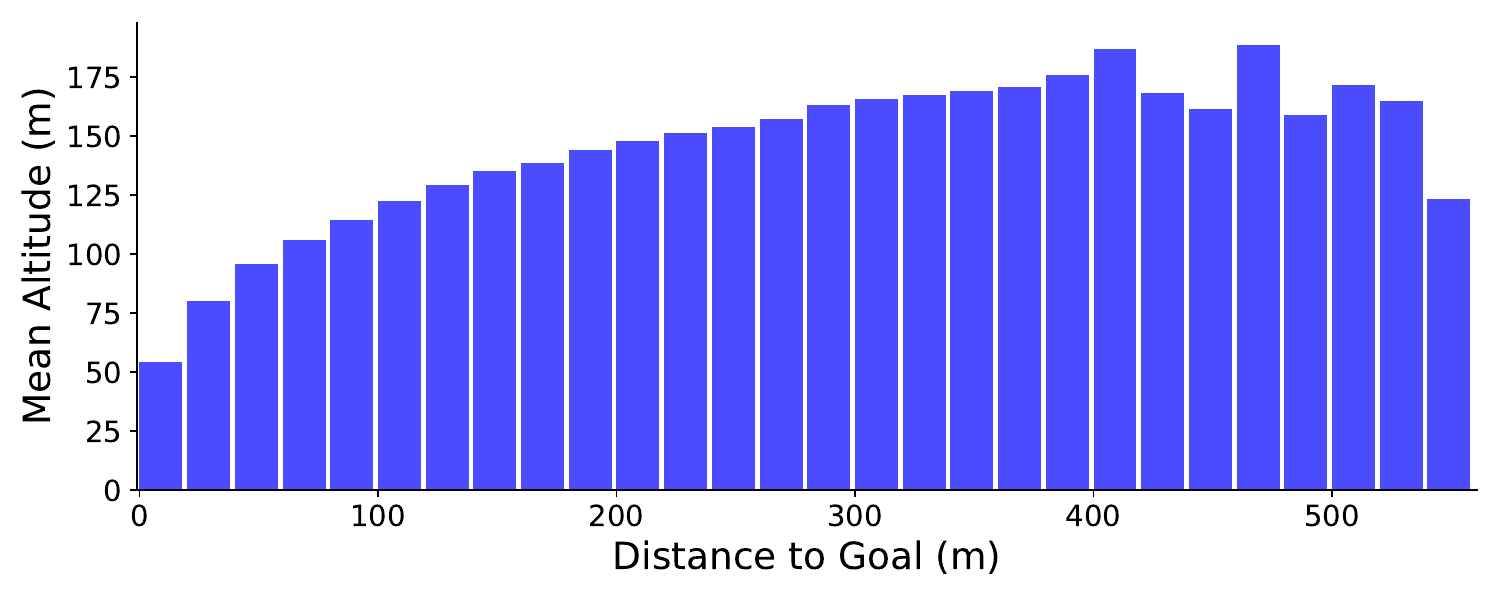}
    \caption{
        Relationship between the distance to goal and the mean altitude of aerial agents.
    }
    \label{fig:mean_altitude}
\end{figure}

\begin{figure}[t]
    \centering
    \includegraphics[width=0.38\textwidth]{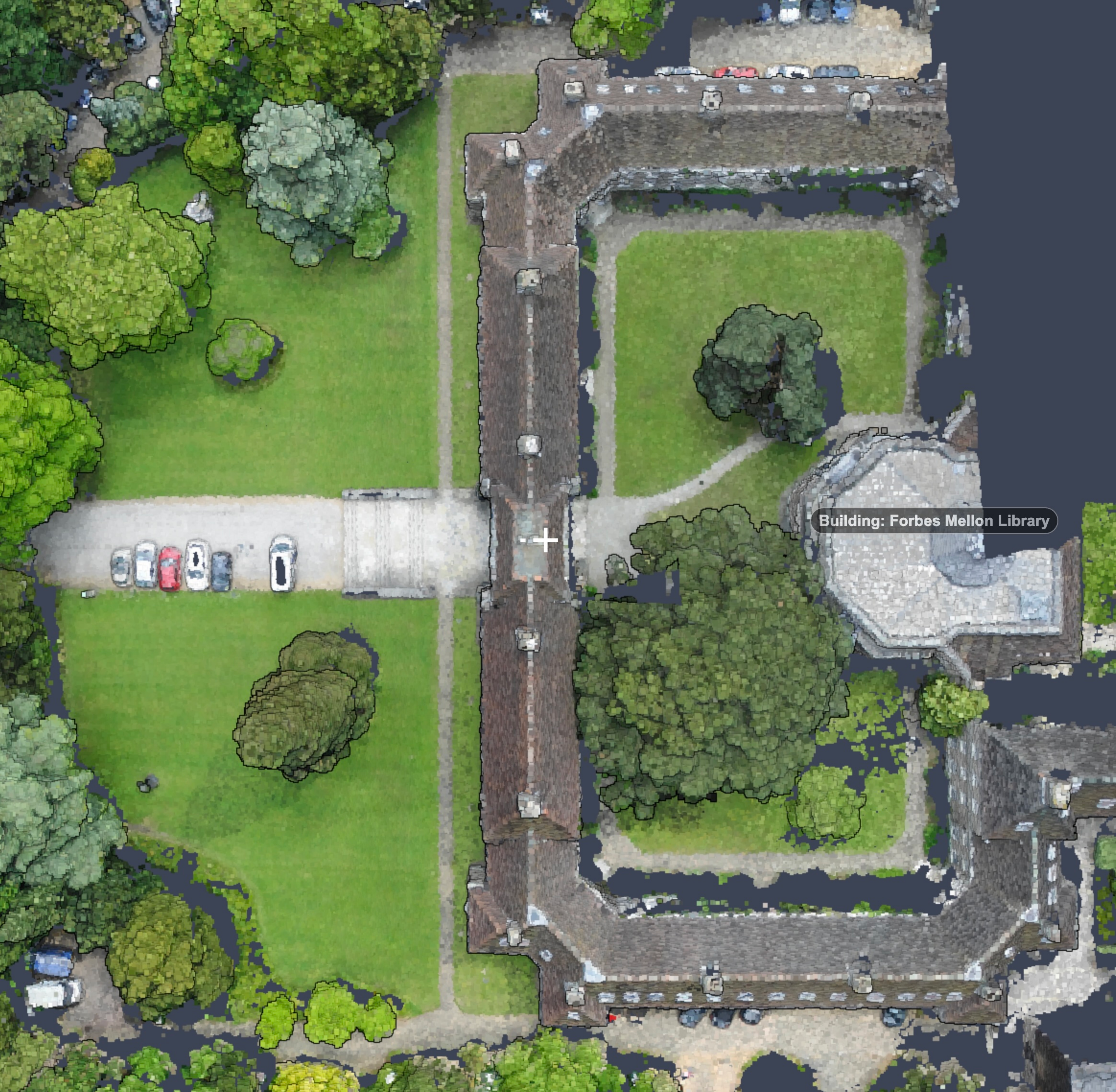}
    \caption{
        Top-down view of the aerial agent at an altitude of 150m, captured via the web interface.
    }
    \label{fig:topdown_view}
\end{figure}

\noindent \textbf{Agent Altitude During Operation.}
We analyze human-operated flights to better understand altitude behavior during navigation tasks. 
Figure~\ref{fig:mean_altitude} shows the mean altitude of human-operated agent trajectories, segmented into 20-meter intervals based on distance from the goal. 
Given that the average 3D altitude is 35.96 meters, this result indicates that most human operators flew above building-level heights, gradually descending as they approached their targets.
In addition, we investigate how clearly ground-level objects are visible at these flight altitudes. 
Figure~\ref{fig:topdown_view} shows a top-down view illustrating that human pilots typically navigate with clear visual access to the target objects. 
Given that the average altitude is above buildings, pilots can effectively identify and target landmarks. 

\vspace{0.5pt}
\noindent \textbf{Human Navigation Strategy.}
In the aerial VLN task, the exploration space is vast, making it crucial to narrow down the search area. 
To address this, our approach mimics the way humans leverage geographic information (landmarks) to reduce the exploration range. As illustrated in Figure~\ref{fig:teaser}, human demonstrations rely on the landmarks mentioned in the description (e.g., {\em Sidney Street}) to navigate toward the landmark’s vicinity. 
Once near the landmark, humans focus their search on the area around it to find the goal object. 
This human strategy enables efficient navigation by focusing efforts around landmarks.

To validate this concept, we analyzed the trajectory data collected in the CityNav dataset, which includes geographic information.
The results indicate that agents passed directly over landmarks 36.3\% of the time in human demonstration (HD) trajectories, compared to 24.6\% in shortest-path (SP) trajectories.
Additionally, we examined whether agents passed within a certain radius of the landmark center. Within 20 meters, 35.5\% of HD trajectories passed near a landmark, compared to 24.0\% for SP. Similarly, at a 40-meter radius, 62.5\% of HD trajectories were near a landmark, compared to 51.9\% for SP.
These results suggest that human pilots tend to navigate closer to landmarks---a strategy that likely contributes to the superior performance of our GSM-based method leveraging human demonstration trajectories, as observed in Table~\ref{tab:necessity} and~\ref{tab:pose_noise}.

\section{Model Details}

\begin{figure}
\centering
\includegraphics[width=\linewidth]{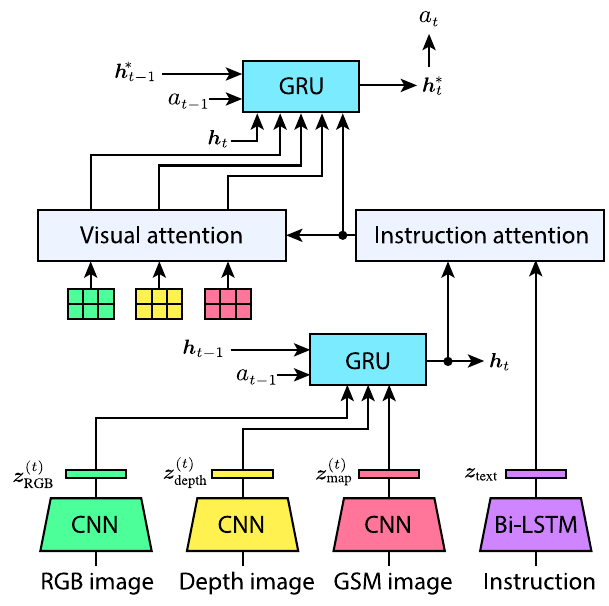}
\caption{Architecture of AerialVLN+GSM.}
\label{fig:aerialvln_arc}
\end{figure}

\subsection{AerialVLN+GSM architecture}
The architecture of AerialVLN+GSM is shown in Figure~\ref{fig:aerialvln_arc}.
It utilizes four input modalities: RGB images, depth images, GSM images, and textual navigation descriptions. For RGB images, a ResNet-50 encoder pre-trained on ImageNet is used. The input size is $224 \times 224$.
For depth images, another ResNet-50 encoder pre-trained on PointGoalNav is used. The input size is $256 \times 256$.
The GSM encoder is a convolutional network consisting of five 2D convolutional layers with channel sizes of (32, 64, 128, 64, 32), kernel size of three, stride of one, and padding of one. Each convolutional layer is followed by a ReLU activation and max-pooling operation. The input size is $224 \times 224$.
The text encoder is implemented using a Bi-LSTM. The output embedding $\bm{h}_{t}^{*}$ is obtained through two GRU modules integrated with description attention and visual attention mechanisms. Specifically, embeddings from the three visual modalities are first fed into the first GRU to produce an intermediate embedding $\bm{h}_{t}$. Subsequently, description attention followed by visual attention for each modality is applied to $\bm{h}_{t}$ and the corresponding modality-specific features. Finally, the second GRU aggregates the outputs from these attention modules to yield the final output embedding $\bm{h}_{t}^{*}$.
The action is predicted from $\bm{h}_{t}^{*}$ through a learnable linear layer.

\subsection{Geographic Semantic Map}
The GSM consists of five categories: current field of view, explored area, landmarks, potential goals, and surrounding objects. These categories are selected because it is essential to understand the spatial relationships between the explored area and objects. The current field of view and explored area are acquired from GNSS coordinates. Specifically, these coordinates are obtained from the CityFlight environment at each time step, and the square area corresponding to the top-down UAV view is marked with a value of one in a binary mask. Landmarks are segments retrieved from OpenStreetMap. For each landmark name, the corresponding segment is retrieved. Potential goals and surrounding objects are detected using an object detector.
We used GroundingDINO~\cite{liu2023grounding} due to its strong performance in open-set object detection.
The detection prompt includes both object categories defined in the SensatUrban dataset and object names extracted from the navigation descriptions (\textit{e.g.}, ``a building with a grey roof'' and ``a red van with black stripes''), to detect object regions from the current RGB image. Before navigation begins, landmark and object names are extracted using a language model (GPT-3.5). The original GSM size corresponds to the smallest 2D map that encompasses the entire 3D scene. Finally, the GSM is resized to 224$\times$224 pixels and provided as input to the model.

\subsection{Training}
All models were trained on a single GeForce RTX 4090 GPU. The Adam optimizer was used for 5 epochs, with an initial learning rate of 5 and a batch size of 12. Cross-entropy loss and an MSE loss, which measures the distance between the goal point and the current position, were employed. For AerialVLN, the step parameter for the look-ahead guidance was set to 10.

\section{Additional Analysis}

\vspace{0.2pt}

\noindent \textbf{Category-level performance.}
We analyze performance at the category level since descriptions can refer to different goal types. Table~\ref{tab:category_performance} shows that AerialVLN+GSM generally delivers the best results, suggesting that integrating the state-of-the-art AerialVLN model with GSM significantly enhances navigation performance at the category level. Although CMA+GSM also shows improvements, it lags behind AerialVLN+GSM, and while Seq2Seq+GSM performs better than its baseline, it remains less effective than the other GSM-enhanced models. Overall, the ground and others categories pose particular challenges for baseline methods, yet GSM integration helps mitigate these difficulties. These findings underscore the value of a geographic semantic map for improving aerial VLN across diverse object categories. Furthermore, the comparison with human performance highlights the gap between aerial agents and human navigation capabilities, with AerialVLN+GSM approaching human-like performance in some metrics while still leaving room for further improvement.

\begin{table}[t]
    \centering
    \small
    \vspace{-10pt}
    \setlength{\tabcolsep}{4pt}
    \renewcommand{\arraystretch}{1.2}
    \begin{tabular}{cccccc}
        \toprule
Category & Method & NE$\downarrow$ & SR$\uparrow$ & OSR$\uparrow$ & SPL$\uparrow$ \\
\midrule
  & Seq2Seq & 244.67 & 1.98 & 8.50 & 1.68 \\      
  & Seq2Seq+GSM & 100.97 & 3.24 & 13.00 & 3.10 \\
  & CMA & 253.16 & 0.76 & 8.73 & 0.72 \\
Building & CMA+GSM & 95.70 & 4.86 & 14.35 & 4.80 \\
  & AerialVLN & 197.51 & 1.71 & 4.00 & 1.61 \\
  & AerialVLN+GSM & \textbf{87.40} & \textbf{6.52} & \textbf{16.91} & \textbf{6.42} \\
  \cmidrule(lr){2-6}
  & Human &  11.3 &  85.64& 93.21 & 57.26 \\
\midrule
  & Seq2Seq & 233.08 & 1.30 & 9.31 & 1.19 \\     
  & Seq2Seq+GSM & 95.78 & 4.11 & 15.44 & 3.96 \\
  & CMA & 239.24 & 0.87 & 11.76 & 0.85 \\
Car & CMA+GSM & 90.99 & 4.98 & 17.75 & 4.95 \\
  & AerialVLN & 164.29 & 2.38 & 4.62 & 2.31 \\
  & AerialVLN+GSM & \textbf{84.78} & \textbf{7.65} & \textbf{18.76} & \textbf{7.52} \\
\cmidrule(lr){2-6}  
  & Human &  6.7 & 95.39& 97.00 & 67.89 \\
\midrule
  & Seq2Seq & 278.82 & 0.59 & 6.93 & 0.59 \\        
  & Seq2Seq+GSM & 88.67 & 3.76 & 14.06 & 3.64 \\
  & CMA & 294.39 & 1.19 & 7.33 & 1.17 \\
Ground & CMA+GSM & 82.31 & 4.16 & 13.47 & 4.06 \\
  & AerialVLN & 208.63 & 0.79 & 2.38 & 0.78 \\
  & AerialVLN+GSM & \textbf{73.05} & \textbf{5.94} & \textbf{19.60} & \textbf{5.87} \\        
\cmidrule(lr){2-6}  
  & Human &  12.0 & 82.40  & 92.42 & 55.64 \\
\midrule
  & Seq2Seq & 245.44 & 0.00 & 3.64 & 0.00 \\        
  & Seq2Seq+GSM & 98.97 & \textbf{3.64} & 10.30 & \textbf{3.64} \\
  & CMA & 232.95 & 0.61 & 9.70 & 0.61 \\
Ohters & CMA+GSM & 89.81 &\textbf{3.64} & 12.73 & 3.60 \\
  & AerialVLN & 182.68 & 1.21 & 2.42 & 1.21 \\
  & AerialVLN+GSM & \textbf{84.65} & \textbf{3.64} & \textbf{21.21} & 3.40 \\
\cmidrule(lr){2-6}  
  & Human & 13.9 & 76.97  & 86.84 & 54.11 \\
        \bottomrule
    \end{tabular}
    \caption{{Performance of each method at the category level.}}
    \label{tab:category_performance}
\end{table}

\paragraph{Disaster scenarios}
Disaster search is one of practical applications as the target's location is unknown. 
We created 2D simulation data for flood scenarios. Table~\ref{tab:flood} summarizes the navigation performance. As shown, all models exhibit reduced performance; however, the effectiveness of GSM remains.
Simulating other types of disasters and more
dynamic scenarios is left for future work.

\newpage
\begin{table}[t]
\vspace{-530pt}
\captionsetup{labelfont={color=black}}
\centering
\small
\setlength{\tabcolsep}{6pt}
\begin{tabular}{lcccc}
\toprule
Method & NE$\downarrow$ & SR$\uparrow$ & OSR$\uparrow$ & SPL$\uparrow$ \\ 
\midrule 
Seq2Seq & {288.5} & {1.38} & {11.58} & {0.69}\\
Seq2Seq+GSM & {98.8} & {3.97} & {14.4} & {2.89}\\
\midrule 
CMA & {273.1} & {0.6} & {9.27} & {0.4}\\
CMA+GSM & {92.5} & {4.61} & {15.63} & {3.47}\\
\midrule 
AerialVLN & {188.6} & {1.46} & {4.65} & {1.38}\\
AerialVLN+GSM & \textbf{84.9} & \textbf{6.80} & \textbf{18.46} & \textbf{6.68}\\
\bottomrule
\end{tabular}
\caption{
Navigation performance under flood inundation conditions (test-unseen).
}
\label{tab:flood}
\end{table}

\end{document}